\documentclass{article}
\usepackage{spconf,amsmath,graphicx,hyperref}
\usepackage{amsfonts}
\usepackage{multirow}
\usepackage{booktabs}
\usepackage{xcolor}
\usepackage[skip=2pt, font=small]{caption}

\title{Uncertainty-Aware 3D Emotional Talking Face Synthesis with Emotion Prior Distillation}
\name{Nanhan Shen and Zhilei Liu\sthanks{Corresponding author (zhileiliu@tju.edu.cn) }}
\address{School of Artificial Intelligence, Tianjin University, Tianjin, China\\
}

\begin{document}
\ninept
\maketitle
\begin{abstract}
Emotional Talking Face synthesis is pivotal in multimedia and signal processing, yet existing 3D methods suffer from two critical challenges: poor audio-vision emotion alignment, manifested as difficult audio emotion extraction and inadequate control over emotional micro-expressions; and a one-size-fits-all multi-view fusion strategy that overlooks uncertainty and feature quality differences, undermining rendering quality.
We propose UA-3DTalk, Uncertainty-Aware 3D Emotional Talking Face Synthesis with emotion prior distillation, which has three core modules: the Prior Extraction module disentangles audio into content-synchronized features for alignment and person-specific complementary features for individualization;
the Emotion Distillation module introduces a multi-modal attention-weighted fusion mechanism and 4D Gaussian encoding with multi-resolution code-books, enabling fine-grained audio emotion extraction and precise control of emotional micro-expressions;
the Uncertainty-based Deformation deploys uncertainty blocks to estimate view-specific aleatoric (input noise) and epistemic (model parameters) uncertainty, realizing adaptive multi-view fusion and incorporating a multi-head decoder for Gaussian primitive optimization to mitigate the limitations of uniform-weight fusion.
Extensive experiments on regular and emotional datasets show UA-3DTalk outperforms state-of-the-art methods like DEGSTalk and EDTalk by $5.2\%$ in E-FID for emotion alignment, $3.1\%$ in SyncC for lip synchronization, and $0.015$ in LPIPS for rendering quality. 
Project page: https://mrask999.github.io/UA-3DTalk/.

\end{abstract}

\section{Introduction}
\label{sec:intro}
Emotional talking face synthesis has become a hot research field in multimedia and signal processing, due to its wide range of utilities\cite{emotkface2025review}.
Current methods mainly suffer from two aspects of technical challenges, each with non-trivial limitations. 
%
The introduction of emotion leads to the first challenge.
Firstly, unlike traditional talking face, exaggerated mouth movements or expressions are common in this scenario, which demands higher synchronization accuracy \cite{flowvq-talker}.
Secondly, audio emotion extraction itself is a challenge, as this is also a hot topic in Speech Emotion Recognition(SER).
Current works mainly rely on reference-based emotion control (e.g., EmotiveTalk\cite{emotive-talk}), which is dependent on high-quality reference video clips and unsatisfactory in pure audio-driven scenarios.
Last but not least, compared to the traditional talking face, this demands stronger control over expressions from audio, which cannot be fully satisfied by the former methods.
%
The second challenge lies in the multi-view feature fusion, which is always overlooked.
Usually, a flawed "one-size-fits-all" strategy is implemented that concatenates all features together \cite{fusionTalking_TOMM}\cite{TKG}\cite{EAMM}, which may generate feature inputs with a length of several hundred dimensions.
In practice, modality quality varies drastically due to occlusions, representation learning issues, and noise; ignoring these differences and inherent uncertainty in multi-view data yields suboptimal results. 
To address these challenges, we propose UA-3DTalk (Uncertainty-Aware 3D Emotional Talking Face Synthesis with Emotion Prior Distillation), a framework that adopts 3D Gaussian Splatting (3DGS) as its rendering backbone and comprises three core modules: Prior Extraction, Emotion Distillation, and Uncertainty-based Deformation.

The implementation of Prior Extraction and Emotion Distillation is fundamental in addressing the issue of poor audio-vision emotion alignment. In Prior Extraction, we disentangled the audio into $f_{exp}$ and $f_{tone}$ as \cite{nerf3d-talker}.
$f_{exp}$ is a content-synchronized feature for audio-vis alignment and $f_{tone}$ is a person-specific complementary feature for individualization, Jointly ensuring synchronization and identity precision.
%
In Emotion Distillation,  we extend the audio window and adopt multi-modal fusion to capture sufficient emotion information without feature dilution.
We extract spectrogram, MFCC, and audio-wav from audio, use the first two as attention keys to re-weight the third for emotion classification and expression prediction~\cite{pd-fgc}, then encode the emotion features via multi-plane gridding~\cite{4dgs} and multi-resolution code-books ~\cite{audio-plane} for precise control of emotional expressions.
%

The Uncertainty-based Deformation module has been proposed as a solution to the problem of suboptimal one-size-fits-all multi-view fusion in existing 3D emotional talking face synthesis methods, uncertainty blocks are implemented for each feature view to jointly generate a state vector for each Gaussian primitive. Unlike previous works such as ~\cite{xie2025}, these blocks separately estimate aleatoric uncertainty (AU) from input noise and epistemic uncertainty (EU) from model parameters and predict view-specific state vectors~\cite{unc-medical, unc2016}.
Adaptive fusion is then implemented based on this information with the rule: higher uncertainty leads to lower view weight. 
The generated state vector is decoded by a multi-head Gaussian deformation decoder~\cite{4dgs} to output modifications of Gaussian primitives. Furthermore, it facilitates the initial systematic integration of uncertainty modeling in the domain of talking face generation. 

In conclusion, main contributions of our work can be summarized as:
\begin{itemize}
    \vspace{-0.1cm}
    \item To address the flawed multi-view fusion issue, we are the first to systematically model aleatoric and epistemic uncertainty in 3D emotional talking face synthesis.
    \vspace{-0.1cm}
    \item We propose an uncertainty-aware adaptive fusion strategy, which dynamically adjusts view weights based on feature quality, prioritizing low-uncertainty information.
    \vspace{-0.1cm}
    \item The Emotion Prior Distillation leverages both the Prior Extraction module and the Emotion Distillation module in a collaborative manner to address the emotion alignment problem.
\end{itemize}

\begin{figure*}[t]
    \setlength{\belowcaptionskip}{-0.5cm}
    \centering
    \includegraphics[width=0.98\linewidth]
    {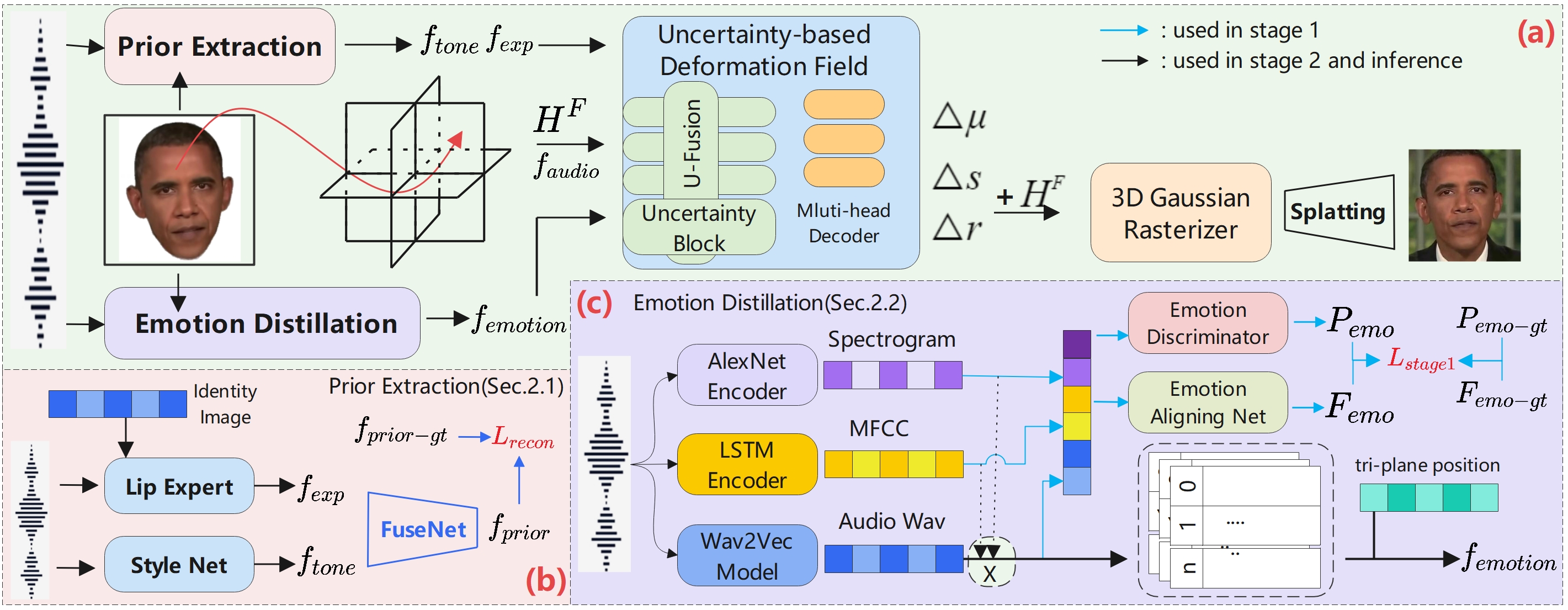}
    \caption{ Overall architecture of the proposed UA-3DTalk framework: (a) End-to-end pipeline of the face branch, integrating Prior Extraction, Emotion Distillation, and Uncertainty-based Deformation for 3D Gaussian primitive adjustment and rendering; (b) Detailed framework of the Prior Extraction module; (c) Technical pipeline of the Emotion Distillation module.}
    \label{pic1}
\end{figure*}

\section{PROPOSED METHOD}
\label{sec:format}
\vspace{-0.2cm}
The architecture of our UA-3DTalk is shown in Fig.~\ref{pic1}, in which 3DGS is adopted as the rendering backbone.
Input image and audio are fed into the Prior Extraction module (Fig.~\ref{pic1}(b)) to generate $f_{exp}$, $f_{tone}$ and into the Emotion Distillation module (Fig.~\ref{pic1}(c)) to obtain $f_{emotion}$.
These features, together with static Gaussian primitives $H$ and the baseline’s simply encoded audio feature $f_{audio}$ ~\cite{TKG}, are input to the Uncertainty-based Deformation module to compute modifications for Gaussian primitives. The Gaussian rasterizer then uses these modified primitives to render dynamic facial frames. Consistent with~\cite{TKG}, we decompose the head into face and mouth branches; the above pipeline corresponds to the face branch (Fig.~\ref{pic1}(a)).

In 3DGS, static gaussian primitives, which model the human head, are defined by a set of point cloud parameters: a 3D center $\mu\in\mathbb{R}^{3}$, a 3D scaling factor $s\in\mathbb{R}^{3}$, a 4D rotation quaternion $r\in\mathbb{R}^{4}$, a scalar opacity value $\alpha\in\mathbb{R}$, and a $Z$-dimensional color feature $c\in\mathbb{R}^{Z}$.
The covariance matrix $\Sigma$ is derived from $s$ and $r$. 
For pixel-level color rendering, the target pixel’s color $C$ is computed by blending contributions from $N$ ordered Gaussian primitives that overlap with the pixel, following Eqs.~\ref{eq01} and~\ref{eq02}.
\vspace{-0.3cm}
\begin{equation}\label{eq01}
C = \sum_{i \in \mathcal{N}} c_i \alpha_i \prod_{j=1}^{i-1}(1 - \alpha_j), 
\qquad \alpha_i = \sigma_i \mathcal{G}_i(\mathbf{x})
\end{equation}
\vspace{-0.3cm}
\begin{equation}\label{eq02}
\mathcal{G}_i(\mathbf{x}) = e^{-\tfrac{1}{2} (\mathbf{x} - \boldsymbol{\mu}_i)^{T} \Sigma_i^{-1} (\mathbf{x} - \boldsymbol{\mu}_i)}
\end{equation}
where $\mathcal{G}_i(\mathbf{x})$ denotes the Gaussian distribution value of the $i$-th primitive at the pixel position $x$, and $\sigma_i$ is a scaling coefficient for the $i$-th primitive's opacity.
To render dynamic facial images, we introduce a deformation field $\delta_i = \{ \Delta \mu_i, \, \Delta s_i, \, \Delta r_i \}$, representing modifications to the Gaussian primitives' position, scaling, and rotation, respectively. 
This deformation field is defined in Eq. \ref{eq03}:
\vspace{-0.2cm}
\begin{equation}\label{eq03}
    \delta_i = Deformation(\mathcal{H}, f_{audio},f_{exp},f_{tone},f_{emotion})
\end{equation}
It is notable that the pipeline of the mouth branch differs slightly from that of the face branch: it excludes the emotional feature $f_{emoiton}$ from the input of the deformation field, and the deformation field only outputs the position modification $\triangle\mu$ for Gaussian primitives. All other components remain consistent with those of the face branch.

\subsection{Prior Extraction}
\label{SecPriorExt}
\vspace{-0.2cm}
To satisfy the elevated requirements of audio-vision synchronization and identity preservation in emotional talking face, the Prior Extraction module is utilized for the generation of highly audio-vision synchronized expression features $f_{exp}$ and person-specific complementary information $f_{tone}$, with the pipeline illustrated in Fig. \ref{pic1}(b). The module operates in three sequential steps: (1) \textbf{Lip Expert Training}: A Lip Expert \cite{lip-expert2020kr} is trained using synchronization loss calculated by a pre-trained SyncNet \cite{lip-expert2020kr}\cite{nerf3d-talker}, with the purpose of generating Lip-wav images to initially establish audio-lip alignment. (2) \textbf{Expression Feature Extraction}: These Lip-wav images are fed into the 3D Prior Extraction sub-module~\cite{sadtalker}, which extracts the expression feature $f_{exp}$, laying the foundation for sustained audio-vision synchronization in subsequent processing. (3) \textbf{Person-Specific Feature Fusion}: StyleNet-based audio encoder~\cite{nerf3d-talker} is introduced to extract the person-specific complementary feature $f_{tone}$ from the input audio. $f_{tone}$ is then fused with $f_{exp}$ to reconstruct 3D facial prior information $f_{gen-exp}$, ensuring the generated features retain individual identity while maintaining synchronization. The training of this part is supervised by reconstruction loss as defined by Eq. \ref{eq04}, with $f_{gt-exp}$ as ground-truth.
\vspace{-0.2cm}
\begin{equation}\label{eq04}
L_{\text{recon}} = \left\| f_{gen-exp} - f_{gt-exp} \right\|^2
\end{equation}

\vspace{-0.2cm}
\subsection{Emotion Distillation}
\label{secEmotionDist}
\vspace{-0.2cm}
The Emotion Distillation module has been developed for the purpose of high-precision audio emotion extraction and strong control of emotional expressions. As shown in Fig.~\ref{pic1}.(c), this module consists of two parts: extraction and encoding.

(1) \textbf{Extraction Component}: To capture sufficient emotion information without feature dilution, we preprocess raw audio into spectrogram, MFCC, and audio-wav features~\cite{wav2vec}, drawing on SER tasks~\cite{ser2022zou}. We extract 2D spectrogram features via AlexNet and time-series MFCC features via LSTM, using these two as attention keys to re-weight audio-wav features into emotion-aware $f_{emo-attn}$, which supports emotion classification and expression feature prediction  for downstream tasks~\cite{pd-fgc, emotive-talk}. The component undergoes two training stages: pre-training on the IEMOCAP database~\cite{iemocap} supervised by cross entropy loss of emotion labels and fine-tuning on person-specific datasets. Fine-tuning loss combines emotion label cross-entropy and L2 prediction loss between predicted  $f_{emo}$ and ground-truth expression features $f_{gt-emo}$, as defined in Eq.\ref{eq_05}.
\vspace{-0.3cm}
\begin{equation}\label{eq_05}
    L_{\text{stage2}} = - \sum_{i=1}^{C} y_i \log(\hat{y}_i) + \left\| f_{emo} - f_{gt-emo} \right\|^2
\end{equation}
Where $C$ is the number of emotion categories, $y_i$ is the one-hot ground-truth label, and $\hat{y}_i$ is the predicted probability. During the post-training phase, the encoder parameters are frozen to preserve extraction capability.

(2) \textbf{Encoding Component}: To handle high-dimensional emotional features incompatible with direct planar gridding~\cite{4dgs}, we adopt multi-resolution code-books~\cite{audio-plane} for discretization and replace concatenation with Hadamard product, which is validated in~\cite{k-planes} to preserve high-dimensional information without dimension explosion. The final encoded feature $ f_{emotion}$ is defined in Eq. \ref{eq06}.
\vspace{-0.15cm}
\begin{equation}\label{eq06}
    f_{emotion} = \bigcup_{(s)}\prod Interp(R^{s}(i,j))\odot f_{emo-attn}^{s}
\end{equation}
where $R$ is plane, $(i,j)\in\{(X,Y),(X,Z),(Y,Z)\}$, $s\in\{1,2,4\}$ denotes multi-resolution scales for code-book discretization, with basic resolution as 64.
These scales enable precise micro-expression control for downstream Gaussian primitive deformation.
%
%

\subsection{Uncertainty-based deformation}
\label{SeqUncertaintyDeform}
\begin{figure}[htb]
    \centering
    \includegraphics[width=0.98\linewidth]
    {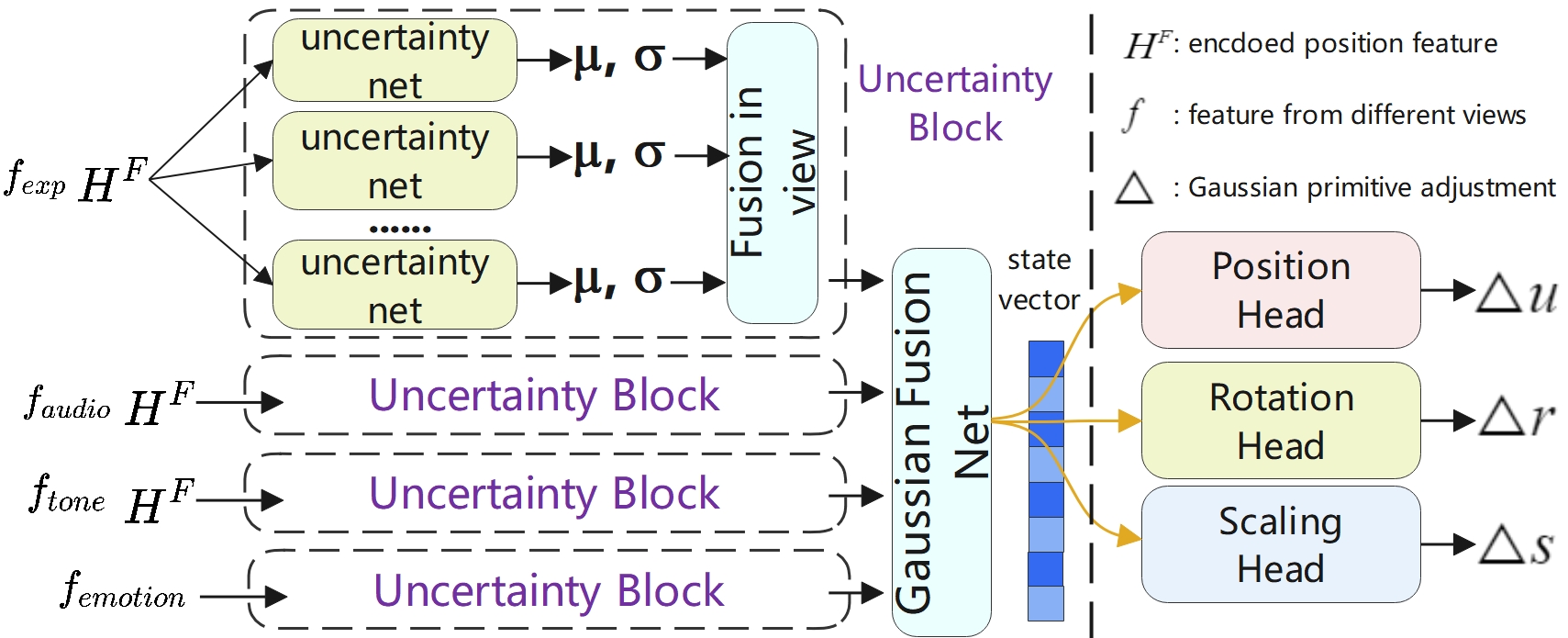}
    \caption{Uncertainty-Based Deformation. (Left) Uncertainty-aware multi-view feature fusion module; (Right) Multi-head Gaussian deformation decoder.} \label{Pic2}
\end{figure}

\vspace{-0.2cm}
The Uncertainty-based Deformation module, composed of feature fusion and decoding components as shown in Fig. \ref{Pic2}, addresses the flawed "one-size-fits-all" multi-view fusion issue by accounting for modality quality differences, aiming to enhance rendering quality and emotional expression precision.

(1)~\textbf{Feature Fusion with Uncertainty Estimation}:
For each feature view, we deploy an uncertainty block consisting of $N$ uncertainty networks. Each network takes Gaussian primitives and the corresponding view’s feature as input (except $f_{emotion}$ input alone), outputting $(\mu, \sigma)$ to represent the predicted distribution of the view’s state vector as shown in Eq.~\ref{eq2}.
\vspace{-0.1cm}
\begin{equation}\label{eq2}
    p(\hat{y}_q \mid x_q, \theta) 
= \mathcal{N}\!\left(\hat{y}; \, \mathbf{F}^{\mu}_{\theta_1}(x), \, \mathbf{F}^{\sigma}_{\theta_2}(x)\right)
\end{equation}
where $\mathbf{F}^{\mu}_{\theta_1}(.)$ and $\mathbf{F}^{\sigma}_{\theta_2}(.)$ are network branches outputting mean $\mu$ and variance $\sigma$, with trainable parameters $\theta_1$ and $\theta_2$.

Following~\cite{unc-medical}, uncertainty includes AU (from input noise) and EU (from model parameters) as defined in Eq.~\ref{eq1}. 
\begin{equation}\label{eq1}
    \mathrm{V}[\hat{y}_q|x_q] 
= \underbrace{ \mathrm{V}_{q^*(\theta)}\!\left[\mathrm{E}(\hat{y}_q|x_q,\theta)\right] }_{\Delta_E[\hat{y}_q]} 
+ \underbrace{ \mathrm{E}_{q^*(\theta)}\!\left[\mathrm{V}(\hat{y}_q|x_q,\theta)\right] }_{\Delta_A[\hat{y}_q]}
\end{equation}
where $\mathrm{E}(\hat{y}_q|x_q,\theta)$ and $\mathrm{V}(\hat{y}_q|x_q,\theta)$ denote the mean and variance of predicted state vector $\hat{y}_{q}$ under posterior predictive distribution $p(\hat{y}|x_{q},\theta)$, with the left representing EU and right representing AU. 

AU is directly estimated by the uncertainty network, while EU is approximated via Monte Carlo sampling: we use $T=10$  uncertainty networks per block, calculating EU as the variance of their $\mu$ outputs. The total uncertainty for each view is formulated in Eq.~\ref{eq3}, with the block’s final output (mean $\hat{\mu}$ and variance $\hat{\sigma}$) defined in Eq.~\ref{eq4} and Eq.~\ref{eq5}.

\vspace{-0.2cm}
\begin{equation}\label{eq3}
{\setlength{\jot}{0pt}
\begin{aligned}
&\hat{\mathbf{V}}[\hat{y}_q \mid x_q] 
= 
\underbrace{\frac{1}{T} \sum_{t=1}^{T} 
\mathbf{F}^{\mu}_{\theta_1^{t}}(x_q)\mathbf{F}^{\mu}_{\theta_1^{t}}(x_q)^{\top} 
- \overline{\mathbf{F}}^{\mu}(x_q)\,\overline{\mathbf{F}}^{\mu}(x_q)^{\top}}_{\hat{\Delta}_{\mathrm{E}}(\hat{y}_q)}\\
&\qquad\qquad\quad+\underbrace{\sum_{t=1}^{T} \mathbf{F}^{\sigma}_{\theta_2^{t}}(x_q)}_{\hat{\Delta}_{\mathrm{A}}(\hat{y}_q)}
\end{aligned}}
\end{equation}
where $\overline{\mathbf{F}}^{\mu}(x_q) 
= \frac{1}{T} \sum_{t=1}^{T} \mathbf{F}^{\mu}_{\theta_1^{t}}(x_q)$ is the mean of output from the $T$ networks. 

\vspace{-0.3cm}
\begin{equation}\label{eq4}
    \hat{\mu}_{y \mid x}= \frac{1}{T} \sum_{t=1}^{T} 
\mu_t\!\left(x; \theta_1^{t}\right) 
\;\xrightarrow[T \to \infty]{}\; 
\mathbb{E}_{q^{*}_{\phi}(y \mid x)}[y]
\end{equation}
\vspace{-0.3cm}
\begin{small}
\begin{equation}\label{eq5}
\hat{\sigma}_{y \mid x} = 
\frac{1}{T} \sum_{t=1}^{T} 
\Big( \sigma_t(x; \theta_2^{t}) 
+ \mu_t(x; \theta_1^{t}) \, \mu_t(x; \theta_1^{t})^{\top} \Big)-\hat{\mu}_{y \mid x} \, \hat{\mu}_{y \mid x}^{\top}
\end{equation}
\end{small}
where $\hat{\mu}$ and $\hat{\sigma}$ are the final output of the current view.

Multi-view features are fused via Gaussian Fusion as defined in Eq.~\ref{eq6}, where weights are dynamically adjusted by uncertainty (lower weight for higher uncertainty).

\vspace{-0.3cm}
\begin{equation}\label{eq6}
    \mu \;=\; \Sigma \sum_{i=1}^{N} \hat{\sigma}_i^{-1} \hat{\mu}_i,
\qquad
\Sigma \;=\; \left( \sum_{i=1}^{N} \hat{\sigma}_i^{-1} \right)^{-1}
\end{equation}
where $N$ is the number of input views, ${\mu}_i$ and ${\sigma}_i$ are the mean and uncertainty of the $i$-th view’s mean and uncertainty.
The final output of feature fusion is predicted state vector $\mu$, which represents audio condition of each Gaussian primitive in emotion space, with real distribution as $q^{*}_{\phi}(y \mid x)$.

(2) \textbf{Gaussian Deformation Decoding}: As shown in Fig.~\ref{Pic2}, the fused state vector is fed into a multi-head Gaussian deformation decoder, replacing conventional simple MLPs, to separately generate modification parameters ($\triangle \mu$, $\triangle r$, $\triangle s$) for Gaussian primitives (position, rotation, scaling). This design aligns with dynamic Gaussian modeling \cite{4dgs}, enabling fine-grained control over each primitive to improve rendering fidelity.

\vspace{-0.4cm}
\subsection{UA-3DTalk Training strategy}
\vspace{-0.2cm}
The training process of our UA-3DTalk framework is divided into two sequential stages, ensuring the independent optimization of core modules and the effective integration of the full rendering pipeline.

\noindent\textbf{Stage 1}: This stage focuses on training the Prior Extraction module and pre-training the Emotion Distillation module. Detailed implementation steps for this stage are specified in their respective sections (Sections~\ref{SecPriorExt} and~\ref{secEmotionDist}). 

\noindent\textbf{Stage 2}: This stage targets the end-to-end optimization of the rendering pipeline, consisting of two key steps:
First, the Uncertainty-based Deformation module is trained separately for the face and mouth branches, supervised by the loss between the ground-truth masked image $\mathcal{I}_{mask}$ and rendered output $\mathcal{I}_{render}$ as defined in Eq.\ref{eq20}. 
Second, fine-tuning is performed to fuse the face and mouth branches, with supervision from the loss between the ground-truth image $\hat{\mathcal{I}}$ and fused output $\mathcal{I}_{fuse}$ as shown in Eq.\ref{eq21}.
\vspace{-0.1cm}
\begin{equation}\label{eq20}
    \mathcal{L}_D = \mathcal{L}_1(\mathcal{I}_{render}, \mathcal{I}_{\text{mask}})
+ \lambda \, \mathcal{L}_{\text{SSIM}}(\mathcal{I}_{render}, \mathcal{I}_{\text{mask}})
\end{equation}
\vspace{-10pt}
\begin{small}
\begin{equation}\label{eq21}
    \mathcal{L}_F = \mathcal{L}_1(\mathcal{I}_{\text{fuse}}, \hat{\mathcal{I}})
+ \lambda \, \mathcal{L}_{\text{SSIM}}(\mathcal{I}_{\text{fuse}}, \hat{\mathcal{I}})
+ \gamma \, \mathcal{L}_{\text{LPIPS}}(\mathcal{I}_{\text{fuse}}, \hat{\mathcal{I}})
\end{equation}
\end{small}
where $\lambda$ and $\gamma$ are both hyperparameters.

\begin{small}
\renewcommand{\arraystretch}{1.1}
\begin{table*}[!t]
    \setlength{\belowcaptionskip}{-0.5cm}
    \centering
    \begin{tabular}{c|cccccc}
    \toprule
         \multirow{2}*{Methods}& \multicolumn{6}{c}{Regular / Emotion}\\
           & LMD($\downarrow$) & PSNR($\uparrow$) & LPIPS($\downarrow$) & SSIM($\uparrow$) & Sync-C($\uparrow$) & E-FID($\downarrow$) \\
        \hline
         StableAvatar \textcolor{gray}{'25} & 4.117 / 7.150 & 18.403 / 19.290
 & 0.258 / 0.228 & 0.480 / 0.619 & 4.421 / 3.972 & 0.546 / 0.430\\
        \hline       
        TalkingGaussian \textcolor{gray}{'24}& 3.018 / 5.934 & 26.943 / 25.533 & 0.045 / \underline{0.096} & \underline{0.906} / 0.892 & 5.011 / 4.886 & 0.089 / 0.356\\
        \hline
        EDTalk \textcolor{gray}{'24}& 3.827 / 6.548 & 25.627 / 18.061 & 0.073 / 0.297 & 0.888 / 0.864 & \textbf{6.173} / \textbf{7.550} & 0.483 / 0.668\\
        \hline
        DEGSTalk \textcolor{gray}{'25}& \textbf{1.960} / \textbf{3.923} & \underline{27.104} / \underline{28.051} & \underline{0.042} / 0.162 & 0.891 / \underline{0.924} & 5.663/5.007 & \underline{0.076} / \underline{0.154}\\
        \hline
        Our & \underline{2.492} / \underline{5.407} & \textbf{28.923} / \textbf{28.408} & \textbf{0.032} / \textbf{0.067} & \textbf{0.928} / \textbf{0.938} & \underline{5.750} / \underline{5.152} & \textbf{0.072} / \textbf{0.145} \\
        \hline
        GT & - & - & - & - & 7.832 / 8.163 & - \\
        \bottomrule 
    \end{tabular}
    \caption{Quantitative results on both datasets. The best methods are in $\textbf{bold}$, and second best are with $\underline{underline}$.}\label{compare1}
\end{table*}
\end{small}

\section{Training and Evaluation}
\subsection{Experimental settings}
\vspace{-0.2cm}
\textbf{Datasets.}\quad To evaluate the effectiveness of our work, experiments are conducted on two types of datasets: both regular talking face dataset \cite{ad-nerf} (including Obama and May subsets) and emotional dataset~\cite{MEAD}(using M003 and M030 subsets). 
For regular dataset, the input video clips are in 25 FPS, with resolution of 256 $\times$ 256.
For emotional dataset, the input video clips are in 30 FPS, with resolution of 512 $\times$ 512.

\noindent\textbf{Preprocessing.}\quad We follow the preprocessing pipeline in~\cite{nerf3d-talker},~\cite{ser2022zou} and initialization settings in~\cite{TKG}. 
To mitigate background clutter interference and camera motion, which is prominent in the MEAD dataset~\cite{MEAD}, we adopt the face window cropping method from~\cite{edtalk}.

\noindent\textbf{Implementation Details.}\quad Our method is implemented on PyTorch.
The structure-related hyperparameters are introduced in each subsection.
In training process, we first train both the face and inside mouth branches for 50,000 iterations in parallel and then jointly fine-tune them for 10,000 iterations.
We use the Adam \cite{adam} and AdamW \cite{adamw} optimizers across all modules, with $\gamma$ and $\lambda$ in the final loss functions set to 0.2 and 0.5.

\noindent\textbf{Evaluation Metrics.}\quad Model performance is evaluated by several established metrics, covering key task requirements: landmark distance (LMD)~\cite{lmd} for facial geometric accuracy, lip synchronization confidence score (Sync-C)~\cite{sync-net}, LPIPS~\cite{lpips} for high-frequency detail fidelity, SSIM~\cite{ssim} for facial structural consistency, and E-FID~\cite{E-FID} for image-level emotion feature alignment.

\vspace{-0.3cm}
\subsection{Comparative Experiments}
\vspace{-0.2cm}

We compare our UA-3DTalk with four SOTA methods, including TalkingGaussian \cite{TKG}, DEGSTalk~\cite{degstalk}, StableAvatar~\cite{sa} and EDTalk \cite{edtalk}. 
Quantitative results in Table~\ref{compare1} demonstrate that our method outperforms them on most metrics across both regular and emotional datasets, with $5.2\%$ higher E-FID  than EDTalk, $3.1\%$ higher SyncC than DEGSTalk, and $0.015$ lower LPIPS than competitors, achieving superior image quality. Notably, DEGSTalk with best LMD uses several ground-truth 3DMM parameters and EDTalk with competitive SyncC relies on reference videos, while our UA-3DTalk excels in pure audio-driven scenarios without such dependencies.

\begin{figure}[h]
    \setlength{\belowcaptionskip}{-0.7cm}
    \centering
    \includegraphics[width=0.98\linewidth]
    {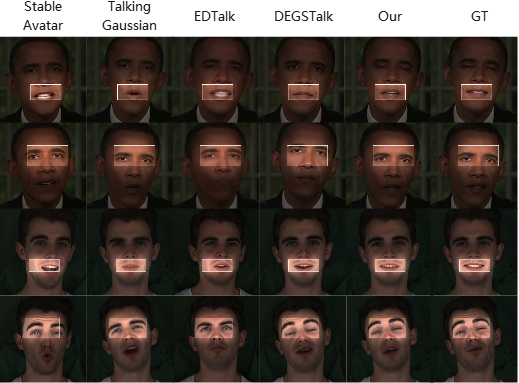}
    \caption{Visual results of the comparative experiments.}
    \label{pic3}
\end{figure}

To further validate our UA-3DTalk’s qualitative performance, we present key frame comparisons on both datasets in Fig. \ref{pic3}. 
We highlight the mouth and eye regions(with consistent positions across rows for focused evaluation.
It can be observed that our method outperforms others in three core aspects: more accurate lip synchronization, more natural eye motion control, and more precise emotion-related expression prediction. The consistency and precision of these dynamic facial regions further confirm the robustness and effectiveness of our proposed framework. 

\vspace{-0.3cm}
\subsection{Ablation Study}
\vspace{-0.2cm}
We conduct an ablation study to quantify the contribution of each core module in UA-3DTalk. 
The tested configurations include: Baseline (without any proposed modules), Prior Extraction (P), Emotion Distillation (E), and Uncertainty-based Deformation (U). 
Results are presented in Table~\ref{compare2}. 
Notably, U cannot be implemented on the Baseline alone, this is because the Baseline only uses single-view features, while U relies on multi-view features for uncertainty estimation. 
Thus, we test U based on the configuration integrated with P to enable comparison with the "w/ P" setting.
\vspace{-0.2cm}
\begin{small}
\renewcommand{\arraystretch}{0.9}
\begin{table}[!h]
\raggedright
\setlength{\tabcolsep}{3pt}
\begin{tabular}{@{}lcccccc@{}} 
\toprule
 & LMD$\downarrow$ & PSNR$\uparrow$ & LPIPS$\downarrow$ & SSIM$\uparrow$ & Sync-C$\uparrow$ & E-FID$\downarrow$ \\
\midrule
Baseline      & 5.934 & 25.53 & 0.096 & 0.892 & 4.886 & 0.356 \\
w/ P     & 5.878 & 28.03 & 0.069 & 0.931 & 4.973 & 0.203 \\
w/ E     & 5.872 & 25.52 & 0.095 & 0.894 & 4.897 & 0.312 \\
w/ P,U   & 5.691 & 28.30 & 0.068 & 0.935 & 5.010 & 0.178 \\
w/ P,E,U & \textbf{5.407} & \textbf{28.40} & \textbf{0.067} & \textbf{0.938} & \textbf{5.152} & \textbf{0.145} \\
\bottomrule
\end{tabular}
\caption{RESULTS OF ABLATION STUDY ON MEAD}
\label{compare2}
\end{table}
\end{small}

\vspace{-0.3cm}
\noindent As shown in Table \ref{compare2}, P significantly boosts performance by generating audio-synchronized action units and expressions, validating its effectiveness. U stably enhances rendering quality without additional information, confirming the validity of our uncertainty-based fusion strategy. Though module E contributes less than module P, it clearly enhances emotion-related micro-expressions and final frames, its effectiveness is fully demonstrated by emotion expression metrics, whether used independently or with module P.
\vspace{-0.4cm}
\section{Conclusion}
\vspace{-0.25cm}

This paper proposes UA-3DTalk to solve two key challenges in 3D emotional talking face synthesis: difficult audio emotion extraction with weak expression control, and flawed "one-size-fits-all" multi-view fusion. It is the first to integrate uncertainty modeling in this field, with its Prior Extraction and Emotion Distillation modules enhancing audio-vision emotion alignment. Experiments show UA-3DTalk achieves state-of-the-art performance, outperforming SOTA methods by $5.2\%$ in E-FID, $3.1\%$ in SyncC, and $0.015$ in LPIPS.

\clearpage

\vspace{-0.4cm}
\section{Acknowledgement}
This work is supported by the Noncommunicable Chronic Diseases-National Science and Technology Major Project (2024ZD0525800).

\begingroup
\fontsize{8pt}{9pt}\selectfont
\bibliographystyle{IEEEbib}
\bibliography{refs}
\endgroup

\end{document}